 \documentclass[sigconf]{acmart}
 \usepackage{xcolor}
\AtBeginDocument{%
  }
    
\acmConference[]{}

\begin{document}

\title{ Functionals in the Clouds:  An abstract architecture of serverless Cloud-Native Apps}

%
%
%
%
%

\author{Stanislaw Ambroszkiewicz  \and
Waldemar Bartyna \and Stanislaw Bylka}
\email{sambrosz@gmail.com, ORCID-0000-0002-8478-6703}


\affiliation{%
  \institution{ University of Siedlce}
  \streetaddress{ 3 Maja 54}
  \country{Poland}
  \postcode{08-110}
}

\renewcommand{\shortauthors}{S. Ambroszkiewicz }

\begin{abstract}
Cloud Native Application CNApp (as a distributed system) is a collection of independent components (micro-services) interacting via communication protocols. This gives rise to present an abstract architecture of CNApp as dynamically re-configurable acyclic directed multi graph where vertices are microservices,  and edges are the protocols. Generic mechanisms for such reconfigurations evidently correspond to higher-level functions (functionals). This implies also internal abstract architecture of microservice as a collection of event-triggered serverless functions (including functions implementing the protocols) that are  composed into event-dependent data-flow graphs, and dynamically reconfigured at the runtime. Again, generic mechanisms for such compositions and reconfigurations correspond to  functionals and higher order type theory like Coq \cite{Coq}. 
Our contribution is strictly theoretical and relies on the abstract architecture of CNApp that is closely related to the calculus of functionals and relations. 
The proposed theoretical approach is an attempt to implement the original idea of programming at the function level postulated by John Backus 1978 \cite{Backus}; the idea that is still waiting to be implemented as a non-von Neumann programming language.
\end{abstract}



\maketitle


\section{Introduction} 

Contemporary {\em  mainframes} are huge data centers comprising  tens or even hundreds of thousands of interconnected physical servers and disk arrays. Usually, they are built according to the cloud architecture where almost everything is virtualized starting with IaaS, PaaS and SaaS.  The next virtualizations  are FaaS (Function as a Service) and BaaS (Backend as a Service) that constitute a relatively new computing paradigm called {\em Serverless}. 
It is based on an execution model in which an application consists of interrelated individual functions and backend services that the cloud provider can manage, scale and execute by automatically and dynamically allocating computing resources.

Developers (of the applications) are charged for—only the compute resources and storage needed to execute a particular piece of code. 


Using FaaS and BaaS, developers can effectively implement micro-services that comprise the applications. Cloud providers are responsible for deploying the code, and for application runtime environment. 

 
There are many serverless platforms provided by main players like  Amazon Web Services, Microsoft Azure, Google Cloud Platform,  IBM Cloud, Alibaba, as well as many others. 
They are trying to persuade developers do create applications mainly out of already existing and proprietary (more or less universal) functions and backend services by composing and orchestrating them into workflows. This gives rise to a new high-level programming paradigm, e.g. Schleier-Smith et al. 2021 \cite{schleier2021serverless}: 
\\ {\em `` 
 This next phase of cloud computing will
change the way programmers work as
dramatically as the first phase changed
how operators work. [...]
new general-purpose serverless abstractions will emerge ...''}
 
The physical servers are based on the von Neumann architecture. The programming languages (for developing functions and backend services) are still of von Neumann style. However, the overall high-level architecture of cloud systems, and the emerging (serverless) programming style (see for example Castro et al. 2019 \cite{castro2019rise}) are not von-Neumann.  This resembles the idea of function-level programming postulated by John Backus 1978 \cite{Backus}.  

The classic functional programming languages (based on lambda calculus, lazy evaluation and term rewriting) like Clojure, Erlang, Scala, Haskell  and many others, are of limited use here. The essence of serverless computing is dynamic orchestration of functions and backend services into workflows where the functions are {\em black boxes} with clearly defined input and output interfaces and functionalities. 

We are going to show that abstract calculus of functionals (higher-order functions) and relations 
(proposed in \cite{CSIT2015}, and then extended in \cite{TO})
 may help to understand and formalize the emerging serverless programming paradigm. Functionals and relations correspond to generic mechanisms to compose and reconfigure serverless functions into complex and sophisticated workflows. They are abstractions that allow to cope with such complexity. 
 
Our contribution is strictly theoretical and relies on the abstract architecture of CNApp that is closely related to the calculus of functionals and relations. From this point of view, the serverless computing paradigm may help to realize the idea of function-level programming postulated by John Backus 1978 \cite{Backus}. 

The von Neumann computer architecture is still the basic abstract model of computation in computers. 

Most of the current programming languages are high-level abstract
isomorphic copies of von Neumann computer architectures.
There is a vicious cycle caused by this isomorphism.
Non-von Neumann computer architectures cannot be developed
because of the lack of widely available and effective non-von
Neumann languages. New languages cannot be created because of
lack of conceptual foundations for non-von Neumann architectures. 

Even though Backus stated this almost 50 years ago, it is still a fundamental truth in IT.

Let us stress again, that the overall high-level architecture of cloud systems, and the emerging (serverless) programming style are not von-Neumann.
 
There are also interesting speculations about the near future of serverless  programming like The Serverless SuperComputer proposed by AWS Lambda creator Tim Wagner in 2019, and The Serverless End Game by Garcia et al. 2021 \cite{garcia2021serverless} predicting virtualization and transparency of computing resources eventually enabling unlimited flexible scaling.

Some related work on workflows and FaaS are as follows. 
%
HyperFlow, Balis 2016 \cite{balis2016hyperflow}, provides an interesting (although explicitly not based on FaaS) model of computation for workflows. The abstraction there is still at the software level. 
Hyperflow by Malawski et al. 2020 \cite{MALAWSKI2020502} was applied to serverless execution of scientific workflows. 

Triggerflow, see Lopez et al. 2020  \cite{lopez2020triggerflow}, is an interesting framework for composing event-based services. Although a formal model of workfow is presented as a Finite State Machine, its architecture and realization are done on the software level.  

Bocci et al. 2021  \cite{bocci2021secure} presented an overview of the existing literature on FaaS orchestrations, and their  executions  environments. 

Ristov at al. 2021 \cite{ristov2021afcl} proposed an abstract function choreography language (AFCL) for serverless workflow specification. The authors claim that AFCL is at a high-level of abstraction. Its basic constructs correspond to the classic programming control-flow and data-flow expressions, and are: if-then-else, switch, sequence, for, while, parallel, and parallelFor. They  are not enough for all possible dynamic  (at  the runtime) reconfigurations of data-flows, and to cope with complexity if the number of functions is large. 
	
All Cloud providers are now offering cloud orchestration and
function composition services, for example,  IBM Composer, Amazon Step Functions, Amazon Express Workfows, Azure Durable Functions, and Google Cloud Composer. All of them are essentially at software level, and do not provide higher-order abstractions to manage the complexity and dynamic re-configurations. 
 
 The necessity of such abstraction is the main idea of Open Application Model (\url{https://oam.dev/}) for defining cloud native apps. It is focused on application rather than container (Docker) or orchestrator (K8s),  and brings modular, extensible, and portable design for modeling application deployment with higher level consistent API. 

To summarize the Introduction, let us cite Kounev et al.  \cite{Kounev} 2023 (Communications of the ACM): 
{\em Even though serverless computing has gained significant attention in industry and academia over the past five years, there is still no consensus about its unique distinguishing characteristics and precise understanding of how these characteristics differ from classical cloud computing.} 
\\
Hence, the fundamental studies (like the one presented in this paper) are still necessary. 

The paper is structured as follows. 
Section \ref{Microservices} provides a short introduction to microservices with an abstract architecture of CNApp  defined as the abstract multi graph of CNApp.  i
In Section \ref{FaaS}, the serverless computing paradigm is roughly described 
along with an abstract view of serverless architecture of microservice as a event-dependent dataflow graph. 
The final Section \ref{TO} contains the summary and conclusions related to John Backus idea \cite{Backus}.




\section{Microservices}
\label{Microservices}

Microservices constitute an architectural pattern in which complex and sophisticated applications are composed from a collection of fine-grained, independent microservices implemented and deployed independently of each other.   
They communicate over a network using lightweight protocols. 
They can be scaled if necessary, e.g. by replication.
Once a collection of such microservices is composed and orchestrated into a dynamic workflow, then it can be deployed on a cloud infrastructure and called a Cloud-Native Application (CNApp for short). 
%
CNApp is a network application. 
It consists of several parts (microservices) that communicate with each other using dedicated, specific protocols implemented on the network protocol stack; for now it is still Internet TCP,UDP/IP.

A general definition of protocol in distributed systems is as follows. 
{\em A protocol defines the format and the order of messages exchanged between two or more communicating entities. The order, types, and the contents of messages depend on the current state of sender. Sending (reception) of a message may cause a change of the state of sender (recipient).} 

Each of the specific protocols of CNApp is based on the client-server model of communication.  It means that the server component is running on a host with a fixed  network address, and is listening on a fixed port waiting for clients.  A client component initializes a communication session to the server; so that the client must know the address and the port number. Usually, the server can communicate with many clients at the same time in separate threads. TCP sockets and UDP sockets are the Internet APIs for such communications.

API (in the context of Cloud computing) is a modern
reincarnation of this very communication protocol in distributed
systems; see, e.g \url{https://aws.amazon.com/what-is/api/} 
and \url{https://thenewstack.io/what-is-api-management/}.

%

HTTP and REST-based protocols (that are commonly used for inter microservice communications) hard-code the format of messages of the real protocols. The HTTP and REST are merely means of transporting messages.
In the case of serverless computing, they are also inefficient due to the additional and high latency overhead.
See also \cite{pemberton2021restless} {\em The RESTless cloud} by Pemberton et al. 2021, for an interesting discussion on serverless communication APIs.  
 
A single microservice may implement and participate in multiple different protocols acting as a client and/or as a server. 

Internal functionality of a microservice is event-driven and composed of serverless functions. Roughly, the incoming messages (as events) trigger serverless functions  (dynamically composed into dataflows) to run, and then to produce outgoing messages. 
This will be explained in details in the next section. 

Microservice scaling by replication and reduction (closing replicas of microservice instances)  forces the microservice to be stateless. If a microservice is statefull, then closing a replica will cause the loss of its state if the state was not stored elsewhere before the closing. 

Actually, the stateless is not a serious restriction, because a stateful microservice (with permanent stored data) may be decomposed into stateless part (where the basic functionality is performed), and bakckend services where the permanent data are stored. 



Hence, a replicate-able microservice can be roughly specified as collection of servers and/or clients of the protocols it participates in, and of its own internal functionality. Permanent data (state) of  microservice are stored in  dedicated backend storing services (BaaS). The internal functionality is composed of stateless elementary functions (FaaS, like the AWS Lambdas), and orchestrated into a dynamic dataflow. 

Usually, communication protocols (at application layer) are defined in an  abstract way independently of their implementation. 

For our propose, {\em  protocol} is defined as two tightly coupled software applications (implementing an abstract protocol): server $S$ and client $P$. Formally, let it be denoted by pair $(P,S)$ with appropriate superscripts and/or subscripts if needed. 

{\em The abstract input} of a microservice can be defined by the servers (of the protocols) it implements:   
$$
IN:=(S_1 ,  S_2 ,  \dots   S_k)
$$ 

{\em The abstract output} of a microservice is defined by the clients (of the protocols) it implements:  
$$
OUT:=(P'_1 ,  P'_2 , \dots   P'_n)
$$ 
Usually, at least one of the clients is for the communication with dedicated backend storing service. 

To omit confusions, server  part  and  client part of protocol will be renamed. 
Components of abstract input will be called {\em abstract sockets}, whereas  components of abstract output will be called {\em abstract plugs}. 

An abstract plug may be connected to an abstract socket if they are of the same type, i.e., they are two complementary sides of the same communication protocol. 
There may be multiple abstract plugs to the same abstract socket. A connection (abstract plug $\to$ abstract socket) is directed meaning that client initializes  a communication session (of the protocol) with server.  

Note that these abstract input and abstract output do not correspond necessarily to data flow. That is, data (messages) may be sent (according to a protocol) out from abstract input (an abstract socket) to an abstract output (an abstract plug). 

Let us formalize the concept described above. 
{\em Microservice}  is defined as 
$$
A := (IN, \mathcal{F}, OUT)
$$
where $IN$ is the abstract input of the microservice, $OUT$ is the abstract output,  and $\mathcal{F}$ is the internal functionality of the microservice to be formally defined in Section \ref{FaaS}. 

Note that abstract plugs (elements of $OUT$) can be dynamically created and/or closed at the runtime. 

Incoming messages via abstract sockets of $IN$ or via abstract plugs of $OUT$  invoke (as events) functions that comprise the internal functionality $\mathcal{F}$ of the microservice. The  functions (in dynamically created dataflow) produce messages to be send out by the abstract output $OUT$ or abstract input $IN$. 

This gives rise to consider a microservice as an {\em abstract function} with abstract input $IN$ and abstract output $OUT$.

\subsection{An abstract architecture of CNApp }
\label{Abstract multi graph of CNApp}

A typical CNApp consists of  the following layers of microservices. 
\begin{enumerate} 
\item
The top layer is for API Gateways. They are entry points of CNApp for users.
Usually, $IN$ of API Gateway has only one element.  It forwards users requests to appropriate microservices. Therefore, API Gateway is supposed to be stateless.
\item 
The second layer consists of regular microservices. Their $IN$ and $OUT$ are not empty. These microservices are also supposed to be stateless. Persistent data (states) of these microservices are stored in backend storage services (BaaS).

\item
The third layer is for backend storage services (BaaS) where all data and files of CNApp are stored. Their $OUT$ is empty. 
\end{enumerate}
%
\begin{figure}
  \begin{centering}
    \includegraphics[width=0.7\linewidth]{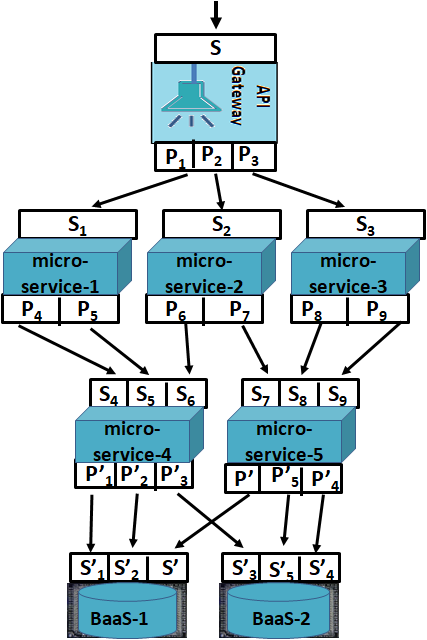}
    \caption{Abstract graph of CNApp - a simple example }
    \label{abstract-graph}
  \end{centering}
\end{figure}

Fig. \ref{abstract-graph} illustrates a CNApp composed of one API Gateway, five stateless regular microservices, and two backend storage services (BaaS).  
Note that the edges denote abstract connections and can also be seen as abstract compositions of microservices within a workflow.

{\em Abstract graph of CNApp} is defined as the following directed labeled multi-graph. 
$$
\mathcal{G}:=( \mathcal{V} , \mathcal{E} )
$$  
where $\mathcal{V}$ and $\mathcal{E}$ denote respectively Vertices and Edges. 
\begin{itemize}
\item Vertices $\mathcal{V}$ is a collection of names of services of CNApp. 
\item 
Edges $\mathcal{E}$ is a collection of labeled edges of the graph. Each edge is of the form:  
$
(A,\ (P,S),\ B)
$
where $A$ and $B$ belong to $\mathcal{V}$,  and $(P,S)$ denotes a protocol. 
That is,  $P$ belongs to $OUT$ of $A$, and $S$ belongs to $IN$ of $B$. Hence, the edges correspond to {\em abstract connections (composition)} between microservices. The direction of edge represents the client-server order of establishing a concrete connection.  There may be multiple edges (abstract connections) between two vertices. 
\end{itemize}

It is postulated that services should be activated only if they are  needed. Also load balancing is necessary by replications  and closing inactive replica. 
For these reasons while running CNApp, some of edges of its abstract graph may be inactive for some time-intervals, i.e. the corresponding protocols sessions are closed or not invoked yet, or can not be invoked for some reasons.  This means dynamic shrinking or expanding of the active part of abstract graph at the runtime. 

%

Netflix uses over 1000 microservices now. Uber now has 4000 or more independent microservices. 
%
If the number of vertices and the number of edges of the abstract graph and its transformations are relatively small (say dozens), it can be done {\em manually} by implementing dedicated management mechanisms.  In general case, when complexity increases dramatically, the design and management of corresponding CNApps  require generic mechanisms. 
%

Since the abstract graphs are abstract compositions of abstract functions, these mechanisms (as higher level abstractions) should be related to abstract functionals, i.e. operations on abstract functions. 

Hence, the calculus of functionals  can serve as a high level abstraction for implementing these generic mechanisms for large and complex CNApps.  
This problem will be addressed in the subsection \ref{Conclusions}. 


\section{Serverless computing paradigm}
\label{FaaS}

{\em Serverless} means that application code (serverless function) is executed on-demand in response to triggers (based on events) that application developers provide in advance.
Usually, serverless functions implement discrete units of application functionality. For an introduction to this subject, see Tozzi 2021  \cite{Tozzi}.

 The serverless computing model includes also Backend as a Service (BaaS) component. It means that the entire backend (database, storage, etc.) of a system is handled independently and offered as a service. 

\subsection{Combining serverless and microservices}

Serverless and microservices are different sorts of technologies. Microservices are a way to design an application, while serverless is a way to run an application (or a part of an application).

Recently, there is an emerging trend to combine serverless and microservices approaches. 
Cloud providers (like Google, Amazon, Microsoft) found ways to bridge the gap between them. 
That is, a microservice can be developed as a collection of interrelated and orchestrated event-driven serverless functions (as a workflow) and stored on the third-party vendor’s infrastructure. It can be done, for example, with Logic Apps (Microsoft), Step Functions (Amazon), or AirFlow workflow engine (Google). They allow assigning triggers to microservices, and combine several functions into a service.  All these workflow engines (as well as others) are at software level, and provide tools to orchestrate (compose) serverless functions into a microservice or a complete application. 
This seems to be the right solution if the number of these functions is relatively small like dozens. 

If, in addition, the workflow requires substantial reconfiguration at runtime, then a higher level of abstraction (beyond software) is necessary to deal with the complexity.

Both serverless functions and microservices (as Service Mesh) require similar approaches to monitoring and management. 

These approaches, like Amazon EventBridge, seem to be over-complicated. Let us present a different approach that can be scaled. 

The crucial question to be answered is:  
{\em What are events in a microservice, and what are they for?} 

Generally, events serve for dynamic (depending on these events) configurations of dataflows composed of  serverless functions (including functions implementing protocols) that comprise functionality of microservice. 

\begin{figure}
  \begin{centering}
    \includegraphics[width=0.3\linewidth]{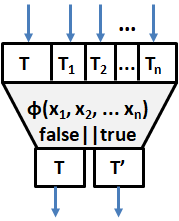}
    \caption{ {\em if-then-else} constructor consisting of formula  $\phi : (T_{1};  T_{2} \dots  ;  T_{n})  \rightarrow boolean $, input type $T$, and two outputs, one of type $T$ and one of type $T'$; the types are the same.  } 
    \label{if-then-else}
  \end{centering}
\vspace*{-0.4cm}
\end{figure}

\begin{figure}
  \begin{centering}
    \includegraphics[width=0.3\linewidth]{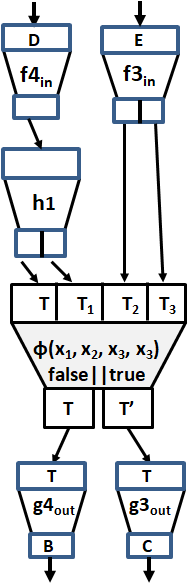}
    \caption{A simple static event-dependent dataflow graph representing a complex  function, say $f$,  of type $(D; E) \to (B || C)$ } 
    \label{simple-dataflow}
  \end{centering}
  \vspace*{-0.3cm}
\end{figure}

\begin{figure}
  \begin{centering}
    \includegraphics[width=0.3\linewidth]{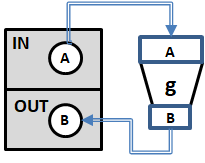}
    \caption{Function $g: A \to B$ is plugged into board where $A$ is socket, and $B$ is plug.  The board may be viewed as the type $A \to B$ } 
    \label{plug-in}
  \end{centering}
  \vspace*{-0.4cm}
\end{figure}

\begin{figure*}
  \begin{centering}
    \includegraphics[width=0.99\linewidth]{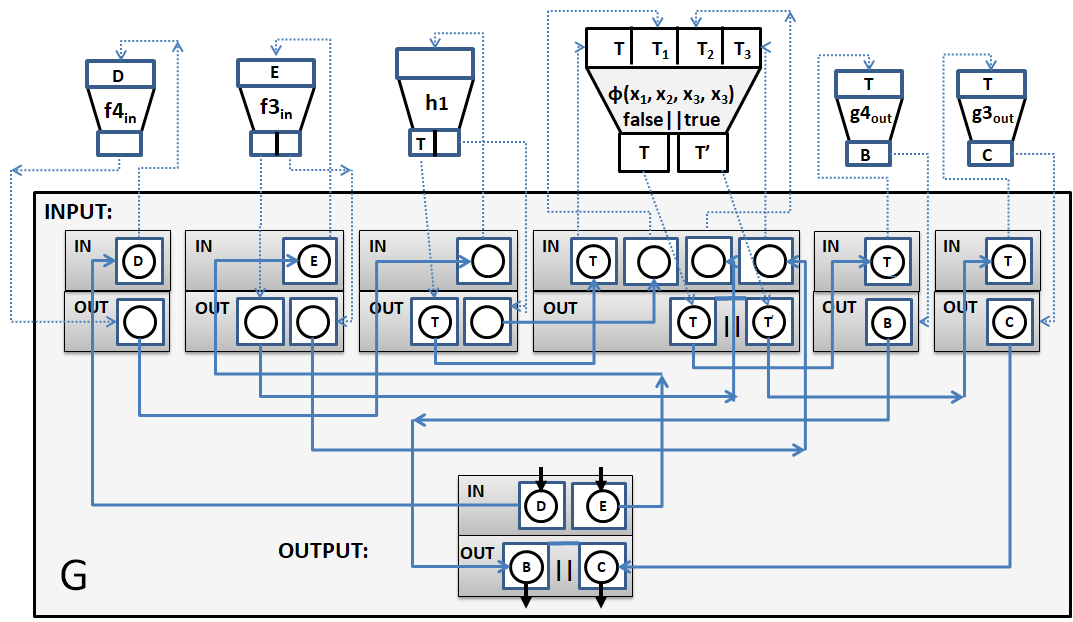}
    \caption{The functional {\bf G} is the complete abstraction of the dataflow graph, i.e. the function $f$ from  Fig. \ref{simple-dataflow} }
    \label{functional}
  \end{centering}
\end{figure*}

\subsection{An abstract view of serverless architecture of microservice} 
\label{An abstract view of serverless architecture of microservice}

Formal foundations of serverless computing is an important research subject in academic community, see for example  Jangda et al. 2019 \cite{jangda2019formal}, Obetz et al. 2020 \cite{obetz2020formalizing}, and Gerasimov 2019 \cite{gerasimov2019dsl}. They are based on classic (based on von Neumann programming style) formal techniques like operational semantics. 

The key concept for the proposed abstract architecture of microservice is {\em relation}. An event may be defined as a relation  evaluated to be either {\em true} or {\em false}. 

Once  proper relations are defined, the constructor {\em if-then-else}  can be used to compose functions into dataflows.  
 Messages, incoming to a microservice, can be considered as {\em coarse events} to be processed in order  to evaluate relations that (as conditions) trigger functions. 

We need a formal framework to define functions and relations.   
Let us start with types, and let $T$ denote a type, with superscripts and/or subscripts if needed.  
We can think of them as data types in programming languages; at this level of abstraction it is not important. 
Let $a:T$ denote that object $a$ is of type $T$. 
%
%
%
Let
$$
f:(T_{1};  T_{2} \dots ;  T_{k}) 
\rightarrow 
(T'_{1} ; T'_{2} \dots ;  T'_{n})
$$
denote function $f$ where $T_{i}$ (for $i=1,2, \dots k$) is the type of its $i$-th input, and  $T'_{j}$ (for $j=1,2, \dots n$) is the type of its $j$-th output. 

Note that $(T_{1};  T_{2} \dots ;  T_{k}) 
\rightarrow 
(T'_{1} ; T'_{2} \dots ;  T'_{n})$ denotes the type of function $f$. 


Function application is denoted 
$
f(e_1, e_2, \dots e_k) 
$
where  $e_i:T_{i}$ for $i=1,2, \dots k$. Result of this application is denoted $(d_1, d_2, \dots d_n)$ where $d_j:T'_{j}$ for $j=1,2, \dots n$; so that \\ 
$
f(e_1, e_2, \dots e_k) = (d_1, d_2, \dots d_n)
$. 

Relation, denoted $\phi$, with  subscripts and/or superscripts if needed,  may be viewed as a boolean-valued function. i.e.
$$
\phi : (T_{1};  T_{2} \dots  ;  T_{k})  
\rightarrow boolean 
$$   
and evaluated as a function. 
Type $boolean$ consists of two logical values: {\em true} and {\em false}. 
In type theory, a special sort {\em Propositions} is needed  for evaluation of relations. For simplicity, the type $boolean$ may serve as a substitute here.  


Functions that process either incoming messages, or produce outgoing messages are defined as follows. 
\begin{itemize}
\item 
 Function  $f_{in}^S$ that processes messages incoming to abstract socket $S$ of protocol $(S,P)$. Let the type of such messages be denoted by $T_{in}^S$. Then,  such a function is of the form 
$$
f_{in}^S: \ T_{in}^S \rightarrow (T_{1} ;  T_{2} \dots  ; T_{n})
$$
\item 
Analogously, for function $f_{in}^P$ that processes messages incoming to abstract plug $P$ of protocol $(S,P)$.   
Let the type of such messages be denoted by $T_{in}^P$. Then,   
$$
f_{in}^P: \ T_{in}^P \rightarrow (T'_{1};  T'_{2} \dots  ; T'_{m})
$$
\item 
Function $g_{out}^S$ that generates messages outgoing from abstract socket $S$ of protocol $(S,P)$. Let the type of such messages be denoted by $T_{out}^S$. 
Then, 
$$
g_{out}^S: \ (T''_{1} ;  T''_{2} \dots ;  T''_{l})
\rightarrow
T_{out}^S 
$$
\item 
Analogously,  function $g_{out}^P$ that produces messages outgoing from abstract plug $P$ of protocol $(S,P)$.   
Let the type of such messages be denoted by $T_{out}^P$. 
Then,  
$$
 g_{out}^P: \ (T'''_{1} ; T'''_{2} \dots  ;  T'''_{k}) \rightarrow T_{out}^P
$$
\end{itemize}

Since it is supposed that the state of a microservice is stored in associated backend services, the protocols (for communications with these backend services) may deliver events informing the microservice on the state change. 
One of  
these protocols can implement one-way communication with an external backend service (clock) to provide (in successive time slots) the current date and time to a microservice in order to synchronize this microservice and the entire CNApp. 

Putting all these pieces together, functionality  $\mathcal{F}$ of microservice can be defined on the basis of: 
\begin{itemize}
\item 
 functions, that process incoming messages, of the form $f_{in}$ with appropriate superscripts corresponding to protocols; 
\item
repository of relations $(\phi_1, \phi_2, \dots ,  \phi_m)$;
\item 
 repository of functions $(h_1, h_2, , \dots , h_l)$;

\item 
 functions, that generate outgoing messages, of the form $g_{out}$ with appropriate superscripts corresponding to protocols. 
\end{itemize}

Functionality  $\mathcal{F}$ of microservice $\mathcal{M}$ can be presented as dataflow graph where initial vertices are functions of the form $f_{in}$ that process incoming messages. Terminal vertices are functions of the form $g_{out}$ that produce outgoing messages. 

Internal vertices of the graph are either functions from repository $(h_1, h_2, , \dots , h_l)$ or constructors of the form {\em if-then-else} to be defined below. 

Directed edges of the graph are just compositions, i.e. links (for data flows)  from an output type of one function to an input type of another function. It means (a partial) composition of these two functions.  The  type of input and the type of output must be the same. 

In Fig. \ref{if-then-else}, the idea of the constructor {\em if-then-else} is presented. Relation 
$
\phi : (T_{1};  T_{2} \dots  ;  T_{n})  
\rightarrow boolean 
$
determines condition  (event) if the values (data) $x_i: T_i$ for $i=1,2, \dots n$ are delivered to the input of $\phi$. If  the relation is evaluated as {\em true}, then the link to output of type $T$ is established.  Otherwise, the link to output of type $T'$ is established. 

A simple example of event-dependent dataflow graph is presented in Fig. \ref{simple-dataflow}. It is important to emphasize that this dataflow is, in fact,  a function (denoted $f$) of type $(D; E) \to (B || C)$, where $B || C$ is the disjoint union of types $B$ and $C$. 

The dataflow graph for a microservice can be complex if the numbers of its functions and its relations  are large. Then, construction, updating functions in the graph, and dynamic reconfiguration may require abstractions related to functionals as it is illustrated in Fig. \ref{plug-in} where the concept of type (as a board) is presented along with pluging the function (as input) into the board. 

If the functional abstraction is applied to the function $f$ from Fig. \ref{simple-dataflow}, then the result is 
the functional {\bf G} presented in Fig. \ref{functional}. 

That functional abstraction consists in replacing each vertex (a particular function), in the dataflow graph presented in Fig. \ref{simple-dataflow}, by the type of this vertex.  

In Fig. \ref{functional}, the functional {\bf G} (the shaded rectangle) consists of: 
\begin{itemize}
\item  INPUT (as the collection of six boards on the top representing respectively the types of the functions: $f4_{in}$, $f3_{in}$, $h1$, $if[\phi ]then-else$, $g4_{out}$, and $g3_{out}$, 
\item OUTPUT as the board on the bottom,  and directed links between them. 
\end{itemize}
Once the functions (on the top):  $f4_{in}$, $f3_{in}$, $h1$, $if[\phi ]then-else$, $g4_{out}$, and $g3_{out}$, are plugged (in the very similar way as in Fig. \ref{plug-in}) into the INPUT boards of {\bf G} (it is shown by dashed directed links), then the result (at the OUTPUT board of ${\bf G}$) is exactly the function $f:(D; E) \to (B || C)$ from Fig. \ref{simple-dataflow}. 

It means that ${\bf G}$,  with the functions plugged in, is {\bf the same} as the dataflow in Fig. \ref{simple-dataflow}. 
That is, $f = {\bf G}(f4_{in}, f3_{in}, h1, if[\phi ]then-else, g4_{out}, g3_{out})$.  

To see so, follow the directed links from socket $D$ and socket $E$ of IN of OUTPUT (via the six INPUT boards on the top, and the plugged functions from the top) to either the plug  $B$ or plug $C$ of OUT of OUTPUT. 
\begin{itemize}
\item 
 from socket $D$ of IN of OUTPUT
 \begin{enumerate}  
\item  to socket $D$ in IN (of board representing the type of  function $f4_{in}$) of INPUT
\item then, following the dotted line, to the socket $D$ of function $f4_{in}$
\item the result of processing by this function is in its plug (output) 
\item from the plug  of $f4_{in}$ (via the dotted line) to OUT of the board representing the type of  function $f4_{in}$
\item then, to the IN of the board (the type of function $h1$)
\item following the dotted line, to the socket of function $h1$
\item the result of processing by this function is in its two plugs that are forwarded via dotted lines to the plugs of the board representing the type of function $h1$
\item from these plugs to the first (from the left) two sockets of the board representing the type of  $if[\phi ]then-else$
\item then, via the dotted lines they are forwarded to the two (from the left)  sockets of the function  $if[\phi ]then-else$. 
\end{enumerate}
\item 
from socket $E$ of IN of OUTPUT
 \begin{enumerate}  
\item  to socket $E$ in IN (of board representing the type of  function $f3_{in}$) of INPUT
\item then, following the dotted line, to the socket $E$ of function $f3_{in}$
\item the result of processing by this function  is in its two plugs (outputs) that are forwarded via the dotted lines to the plugs  (OUT) of the board representing the type of function $f3_{in}$
\item from these two plugs to the two plugs of the board representing the type of $if[\phi ]then-else$
\item  then, via the dotted lines, they are forwarded to the two (from the right)  sockets of the function  $if[\phi ]then-else$. 
\end{enumerate}
\item 
the input data from these four sockets of the function $if[\phi ]then-else$ is processed by this function
\item the result of this processing is either in the first plug denoted T, or in the second plug denoted T'; note that the type of plug T and the type of plug T' are the same  
\item if the result is in the plug T of $if[\phi ]then-else$, 
\begin{enumerate}
\item it is send to the first (from the left) plug of the board representing the type of $if[\phi ]then-else$, 
\item and then forwarded  to the socket of the board representing the type of function  $g4_{out}$
\item then, via the dotted line, to the socket T of $g4_{out}$ where it is processed by this function
\item the result (output) is in the plug B, and then is forwarded (via the dotted line) to the plug of the board representing the type  of $g4_{out}$
\item finally, the result is sent to the plug B of the board OUTPUT. 
\end{enumerate}
\item otherwise. i.e. if the result is in the plug T' of $if[\phi ]then-else$, 
\begin{enumerate}
\item it is send to the first (from the right) plug of the board representing the type of $if[\phi ]then-else$, 
\item and then forwarded to the socket of the board representing the type of function  $g3_{out}$
\item then, via the dotted line, to the socket T of $g3_{out}$ where it is processed by this function
\item the result (output) is in the plug C, and then is forwarded (via the dotted line) to the plug of the board representing the type  of $g3_{out}$
\item finally, the result is sent to the plug C of the board OUTPUT. 
\end{enumerate}
\end{itemize}
The dataflow described above is exactly the dataflow of the graph of Fig. \ref{simple-dataflow}.  

Hence, the functional ${\bf G}$ represents the complete connectivity of the dataflow graph of Fig. \ref{simple-dataflow} abstracted from the concrete functions placed in the vertices of the graph. These functions can be replaced by different functions if only their types are appropriate. Then, the resulting dataflow graph is syntactically correct, and can be executed if only data (incoming messages) are delivered to its inputs.  

Hence, ${\bf G}$  with the functions plugged in (as in Fig. \ref{functional}) is {\bf the same} as the dataflow in Fig. \ref{simple-dataflow}. 
It means that ${\bf G}$ expresses the essence of the dataflow graph connectivity, i.e. any vertex (concrete function) of the graph may be changed by any function of the same type as the type of the vertex. 


\subsection{Conclusions}
\label{Conclusions}

{\bf \color{blue} Note that an analogous abstraction to the functional can be performed for the abstract multigraph of CNApp from Fig. \ref{abstract-graph}. }

{\bf  \color{red} Hence, a functional is an abstraction consisting of boards (representing types) and connections representing dataflows. In general case, such boards and connections can be dynamically configured depending on flow of concrete data. }

There is a similar approach CIEL (a bit old) Murray et al. 2011 \cite{murray2011ciel} where there are also similar  data-dependent control flow constructors, and iterations as dynamic unfolding of  acyclic dataflow graphs. In our approach we introduce much more, i.e. functionals as generic mechanisms for transformations of the graphs at the runtime. These mechanisms include higher order primitive recursion schema, dependent types, and much more. 

Dynamic creation and/or closing abstract plugs of $OUT$ of a microservice transforms its dataflow graph at the runtime by adding or removing appropriate functions.  New  connections (initialized by outside clients, i.e. by abstract plugs of another microservices) to abstract sockets  of the microservice expand the dataflow with appropriate functions. 
Closing connections of some of the outside clients results in removing corresponding functions. 

Therefore, in the general case, a running microservice can dynamically transform its data flow graph.
The graph transformations may be caused also by the input from backend services where the state of microservice can be stored. The input may dynamically change the basic functionality of the microservice at the runtime. 
Dynamic transformations of dataflow graphs  can also be viewed as functionals. Calculus of functionals and relations can be seen as a general framework for constructing dynamic dataflow graphs.

\section{Summary} 
\label{TO}

The proposed abstract architecture of {\em serverless} Cloud-Native Apps consists of two layers of abstraction. 

{\bf Top level} concerns the so called Service Mesh. The abstract view of CNApp (consisting of interacting microservices) is presented in Section \ref{Abstract multi graph of CNApp} as abstract directed multigraph of CNApp  for orchestration of its microservices; see the companion paper \cite{SSMMP} for details.  
{\bf Bottom level} is about the functionality of microservice. Dynamic event-driven dataflow graph (from the previous Section) is proposed as an abstract view of such functionality. 

If the proposed abstract architecture is of some interest as a theoretical approach, then the following challenges emerge. 
\begin{itemize}
\item
Repository of standard generic functions. 
\item
Repository of standard communication protocols, and corresponding functions implementing the protocols.  
\item
Repository of standard generic relations.  
\item
Repository of standardized microservices build on the basis of the above repositories. 
\item 
Platforms and tools for developing these repositories. 
\item
Tools and generic mechanisms for constructing the graphs and transforming them. 
\item 
Managing the complexity of the internal structure of a microservice if the number of its functions is large.
\item
Managing complexity of  CNApp if the number of its microservices in large. 
\end{itemize}
\noindent
Calculus of functionals and relations provides abstractions for reducing complexity of:
 \begin{itemize}
 \item
 Construction of the dataflow graphs of microservices, and the abstract graphs of CNApps.   
 \item
 Dynamic reconfiguration and transformations of the graphs at  the runtime.  
\end{itemize}

It seems that for the both levels of abstraction,  a calculus of functionals and relations (like the one proposed in \cite{TO} and \cite{CSIT2015})  can serve as a theoretical foundations for an abstract architecture of serverless CNApps. 

The above theoretical framework strongly relates to the original idea of function-level programming postulated by John Backus 1978 \cite{Backus}.
The idea that is still waiting to be implemented as an information technology. 

\bibliographystyle{ACM-Reference-Format}
\bibliography{FC}
\end{document}